# Employing fuzzy intervals and loop-based methodology for designing structural signature: an application to symbol recognition

MM.Luqman[*], M. Delalandre[+], T. Brouard[*], JY.Ramel[*] and J. Lladós[+]

[*] Université François Rabelais de Tours,Laboratoire d'Informatique (EA 2101),37200 Tours – France
E-mail: muhammadmuzzamil.luqman@etu.univ-tours.fr, {brouard, ramel}@univ-tours.fr

[+] Centre de Visió per Computador, 08193 Bellaterra (Barcelona) – Spain
E-mail: {mathieu, josep}@cvc.uab.es

**Abstract**

Motivation of our work is to present a new methodology for symbol recognition. We support structural methods for representing visual associations in graphic documents. The proposed method employs a structural approach for symbol representation and a statistical classifier for recognition. We vectorize a graphic symbol, encode its topological and geometrical information by an ARG and compute a signature from this structural graph. To address the sensitivity of structural representations to deformations and degradations, we use data adapted fuzzy intervals while computing structural signature. The joint probability distribution of signatures is encoded by a Bayesian network. This network in fact serves as a mechanism for pruning irrelevant features and choosing a subset of interesting features from structural signatures, for underlying symbol set. Finally we deploy the Bayesian network in supervised learning scenario for recognizing query symbols. We have evaluated the robustness of our method against noise, on synthetically deformed and degraded images of pre-segmented 2D architectural and electronic symbols from GREC databases and have obtained encouraging recognition rates. A second set of experimentation was carried out for evaluating the performance of our method against context noise i.e. symbols cropped from complete documents. The results support the use of our signature by a symbol spotting system.

*Keywords:* symbol recognition, fuzzy interval, structural signature, Bayesian network.

## 1   Introduction and related works

Graphics recognition is a subfield of document image analysis and it deals with graphic entities that appear in document images. As pointed out by Lladós and Sánchez in [1]: documents from electronics, engineering, music, architecture and various other fields use domain-dependent graphic notations which are based on particular alphabets of symbols. These industries have a rich heritage of hand-drawn documents and because of high demands of application domains, overtime symbol recognition is becoming core goal of automatic image analysis systems. Some typical applications of symbol recognition include hand-drawn based user interfaces, backward conversion from raster images to CAD, content based retrieval from graphic document databases and browsing of graphic documents. A detailed discussion on application domains is in [2, 3] and a quick historical overview of work on graphic symbol recognition is given by Tombre et al. [4].

Graphic symbol recognition is generally approached by structural methods of pattern recognition which normally use graph based representations and thus inherit the various advantages associated with these representations. These methods, for example [5, 6] and the methods mentioned in [1], then employ graph matching or graph comparison techniques for symbol recognition. Graph matching and graph comparison are time consuming tasks and they limit the ability of these systems to scale to large number of symbol models. Moreover, structural methods generally require in-depth domain knowledge and this hinders the possibility of having a generalized system of symbol recognition. Another approach for graphic symbol recognition is use of statistical methods of pattern recognition. These methods represent graphic symbol by feature vector or signature (we use these terms interchangeably) and use a statistical classifier for symbol recognition. The use of signatures and statistical classifiers allows designing of fast and efficient systems which are sufficiently scalable and domain independent. A state of the art for various methods that employ different structural or statistical approaches for graphic symbol recognition is in [7].

The rest of paper is organized as follows: section 2 gives a general description of our method and section 3 is devoted to detailed description of each part of the proposed system. Experimental results are presented in section 4 and we finally present some concluding remarks with future direction of work in section 5.

## 2    Proposed method

### 2.1 A combination of structural and statistical approaches

We have approached the problem of graphic symbol recognition by employing a structural method for symbol representation and a statistical classifier for recognition. In this paper we take forward the work of Qureshi et al. [8]. They vectorize a graphic symbol, construct its ARG[1] and compute a structural signature for it. They call it a graphical signature or G-Signature. For classification of query symbol they use nearest neighbor rule with Euclidian distance as measure of dissimilarity. The structural signature is discriminant in case of hand-drawn or vectorial deformations and has been shown invariant of rotation and scaling. We argue that the computation of Euclidian distance in a brute force manner (between query symbol and each prototype in training set) limits this system to scale to large number of symbol models or to be used by real time applications. The system is based on vectorization and faces a high degree of uncertainty as the level of noise and deformation increase. In our system we use structural signature with a statistical classifier. We have selected Bayesian networks for dealing with uncertainty in symbol signatures and propose to use overlapping or fuzzy intervals instead of rigid boundaries for computing features from ARG of symbols. We deal only with linear graphic symbols in this work i.e. symbols that consist of only straight lines and arcs. This gives us a chance to optimize the structural signature for these types of symbols. The signature is given in Figure 2 and it is discussed in section 3.2.

### 2.2 Bayesian networks

Bayesian networks are probabilistic graphical models and are represented by their structure and parameters. Structure is given by a directed acyclic graph and it encodes the dependency relationships between domain variables whereas parameters of the network are conditional probability distributions which are associated with its nodes. A Bayesian network, like other probabilistic graphical models, encodes the joint probability distribution of a set of random variables, and could be used to answer all possible inference queries on these variables. A humble introduction to Bayesian networks is in [9, 10].

Bayesian networks have already been applied successfully to a large number of problems in machine learning and pattern recognition and are well known for their power and potential of making valid predictions under uncertain situations. But in our knowledge there are only a few methods which use Bayesian networks for graphic symbol recognition. Recently Barrat et al. [11] have used the naïve Bayes classifier in a 'pure' statistical manner for graphic symbol recognition. Their system uses three shape descriptors: *Generic Fourier Descriptor, Zernike descriptor and R-Signature 1D,* and applies dimensionality reduction for extracting the most relevant and discriminating features to formulate a feature vector. This reduces the length of their feature vector and eventually the number of variables (nodes) in network. The naïve Bayes classifier is a powerful Bayesian classifier but it assumes a strong independence relationship among attributes given the class variable. We believe that the power of Bayesian networks is not fully explored; as instead of using pre-defined dependency relationships we can obtain a better Bayesian network classifier if we find dependencies between all variable pairs from underlying data. This will help us to ignore irrelevant variables and exploit the variables that are interesting for discriminating symbols in underlying symbol set (section 3.3 and section 3.4).

---

[1] Attributed Relational Graph

## 2.3 Originality of our approach

Our method is an original adaptation of Bayesian network learning for the problem of graphic symbol recognition. We use a structural signature for symbol representation; the signature is computed from the ARG of graphic symbol and is composed of geometric and topologic characteristics of the structure of symbol. We use fuzzy intervals (overlapping intervals) for computing noise sensitive features of our signature and this increases the ability of our signature to resist against irregularities that may be introduced in the shape of symbol by deformations and degradations. We employ a Bayesian network for symbol recognition. This network is learned from underlying training data by using the quite recently proposed genetic algorithms for Bayesian network learning by Delaplace et al. [12]. A query symbol is classified by using Bayesian probabilistic inference (on encoded joint probability distribution). We have selected the features in signature very carefully to best suit them to linear graphic symbols and to restrict their number to minimum; as Bayesian network algorithms are known to perform better for a smaller number of nodes. The use of structural signature makes our system independent of application domains and it could be used for all types of 2D linear graphic symbols. Also, relatively basic computations are involved for recognizing a query symbol which enables our system to respond in real time and it could be used for instance as a pre-processing step of a traditional symbol recognition method or for indexation and browsing of graphic documents.

# 3    Detailed description

Cordella and Vento [7] have remarked that a graphics recognition system can be looked upon as working in three phases: representation phase, description phase and classification phase. In this section we describe our system in light of these phases.

## 3.1 Representation phase

This important phase of our system is based on method proposed by Qureshi et al. in [8] and is summarized in Figure 1. The representation phase extracts the topological and geometric details about structure of symbol and represents them by an ARG. The symbol is vectorized and is represented by a set of primitives (labels 1, 2, 3, 4 in Figure 1). These primitives become nodes and topological relations between them become arcs in ARG. Nodes have *'relative length'* (normalized between *0* and *1*) and *'primitive-type'* (*Vector* for filled regions of shape and *Quadrilateral* for thin regions) as attributes; whereas arcs of the graph have *'connection-type'* (*L, X, T, P, S*) and *'relative angle'* (normalized between *0°* and *90°*) as attributes.

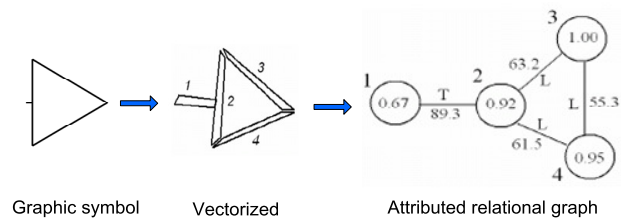

Graphic symbol    Vectorized    Attributed relational graph

Figure 1. Representation phase.

## 3.2 Description phase (fuzziness of signature)

This phase concerns the extraction of features from ARG and their use for computing a structural signature. We introduce overlapping intervals to improve the structural signature proposed by Qureshi et al. [8]; in order to enable it to take care of irregularities in signature. These irregularities are in fact a result of degradations and deformations in shape of symbol and are caused by noise. Figure 2 presents our proposed structural signature for graphic symbol. Our motivation for choosing structural features for signature is to exploit their ability to identify symbols in context [13].

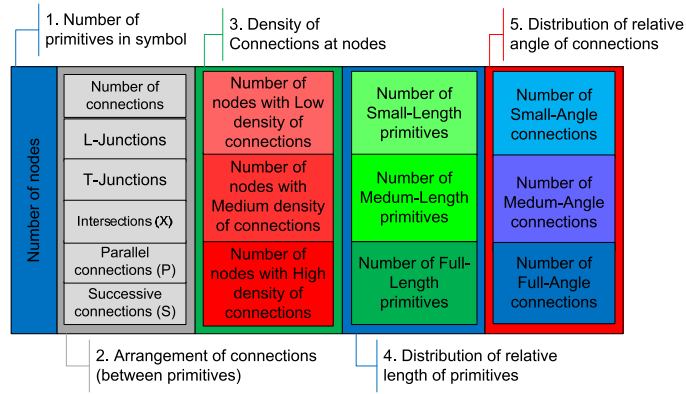

Figure 2. Structural signature for graphic symbol.

Group 1 & 2 encode the size of symbol and arrangement of its primitive components respectively. These features discriminate between symbols of different sizes and also between symbols of same size but with a different arrangement of primitives. Group 3 encodes the density of connections for nodes. This group discriminates between symbols that have similar number of primitives with a similar arrangement but different density of connections at its nodes. Group 4 & 5 exploits the attributes of ARG and encodes details of length and angle attributes. These groups complement the criteria (of Groups 1, 2 & 3) for discriminating between symbols that belong to different classes.

The computation of features in Group 1 & 2 is straightforward and is achieved by counting the relevant information in ARG of graphic symbol. For features in Group 3 we first compute a list of connection densities of all nodes of all ARG (of symbols in underlying symbol set). And then use this list of connection-density counts for finding connection-density intervals for computing feature in Group 3 of structural signature. We use a histogram based binning technique from [14] for this purpose. The technique is originally proposed for discretization of continuous data and is based on use of Akaike Information Criterion (AIC). It starts with an initial m-bin histogram of data and finds optimal number of bins for underlying data. Two adjacent bins are merged using an AIC-based cost function; until the difference between AIC-before-merge and AIC-after-merge becomes negative. We arrange these bins in overlapping fashion (fuzzy approach) and use them as intervals for computing number of nodes lying in different connection-density intervals. This gives us a distribution of nodes in structural graph with low, medium and high density of connections, which we use as features of our signature.

Group 4 (& 5) is computed by dividing relative length (and relative angle) in three overlapping intervals, as shown in Figure 3 (and Figure 4). The overlapping intervals (fuzzy approach) handle the irregularities caused by distortions and degradations, and ensure that these irregularities do not affect the signature.

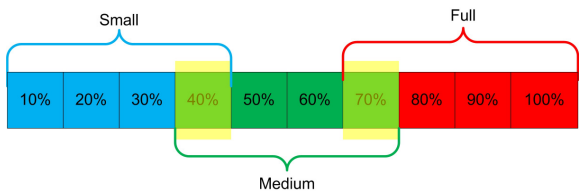
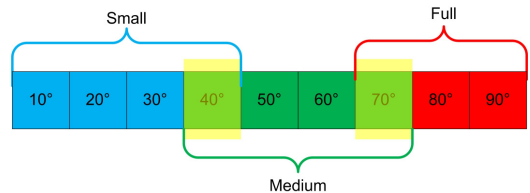

Figure 3. Intervals for computing number of small, medium and full length primitives.

Figure 4. Intervals for computing number of small, medium and full angle connections.

### 3.3 Learning phase

After representing the symbols in learning set by ARG and describing them by structural signatures, the signatures are discretized [14] and are used for learning a Bayesian network. We discretize each feature variable (of signature) separately and independently of others. The class labels are chosen intelligently in order to avoid the need of any discretization for them. The discretization of '*number of nodes*' and '*number of arcs*' achieves a comparison of similarity of symbols (instead of strict comparison of exact feature values). This discretization step ensures that the features in signature of query symbol will look for symbols whose number of nodes and arcs lie in same intervals as that of the query symbol.

The Bayesian network is learned in two steps. First we learn the structure of the network by genetic algorithms proposed by Delaplace et al. [12]. Each feature in signature becomes a node of network. The goal of structure learning stage is to find the best network structure from underlying data which contains all possible dependency relationships between all variable pairs. And the structure of the learned network depicts the dependency relationships between different features in signature. Figure 5 shows one of the learned structures from our experiments.

The second step is learning of parameters of network; which are conditional probability distributions (P (node$_i$|parents$_i$)) associated to nodes of the network and which quantify the dependency relationships between nodes. The network parameters are obtained by maximum likelihood estimation (MLE); which is a robust parameter estimation technique and assigns the most likely parameter values to best describe a given distribution of data. We avoid null probabilities by using Dirichlet priors with MLE. The learned Bayesian network encodes joint probability distribution of the symbol signatures.

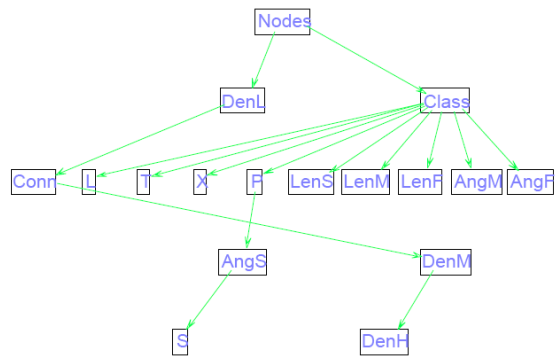

Figure 5. A Bayesian network structure after learning step; each node corresponds to a feature variable.

The conditional independence property of Bayesian networks helps us to ignore irrelevant features (in structural signature) for an underlying symbol set. This property states that a node is conditionally independent of its non-descendents given its immediate parents [9]. Conditional independence of a node in Bayesian network is fully exploited during probabilistic inference (see section 3.4) and thus helps us to ignore irrelevant features for an underlying symbol set while computing posterior probabilities for different symbol classes (see section 3.4).

### 3.4 Classification phase (graphic symbol recognition)

For recognizing a query symbol we use Bayesian probabilistic inference on the encoded joint probability distribution. This is achieved by using junction tree inference engine which is the most popular exact inference engine for Bayesian network probabilistic inference and is available in [14]. The inference engine propagates the evidence (signature of query symbol) in network and computes posterior probability for each symbol class. Equation 1 gives Bayes rule in terms of our system. It states that posterior probability or probability of a symbol class 'c$_i$' given a query signature 'evidence e' is computed from likelihood (probability of 'e' given 'c$_i$'), prior probability of 'c$_i$' and marginal likelihood (prior probability of 'e').

$$P(c_i | e) = \frac{P(e, c_i)}{P(e)} = \frac{P(e | c_i) \times P(c_i)}{P(e)} \quad (1)$$

*Where,*

$$e = f1, f2, ..., f21$$
$$P(e) = P(e, c_i) = \sum_{i=1}^{k} P(e | c_i) \times P(c_i)$$

We compute the posterior probabilities for all symbol classes and assign query signature to class which maximizes posterior probability i.e. which has highest posterior probability for the given query symbol.

## 4     Experimental results

### 4.1 Experimentation – symbols with vectorial and binary noise

We have experimented with synthetically generated 2D symbols of models collected from databases of GREC symbol recognition contest [15]. There are a total of 150 models in GREC databases (the models are available on ISRC2005 website[2]). These models belong to the domains of architecture and engineering, and are composed of only straight lines and arcs. In order to get a true picture of the performance of our proposed method on this database, we have randomly selected subsets with 100, 75, 50 and 20 different classes and generated our learning and test sets for each of these subsets. For each class the perfect symbol (the model) along with its 36 rotated and 12 scaled examples was used for learning; as the features have already been shown invariant to scaling and rotation [8] and because of the fact that generally Bayesian network learning algorithms perform better on datasets that contain quite a good number of examples. The system has been tested for its scalability on clean symbols (rotated/scaled), various levels of vectorial deformations and for binary degradations of GREC symbol recognition contest (Figure 6 and Figure 7). Each test dataset was composed of 10 query symbols for each class.

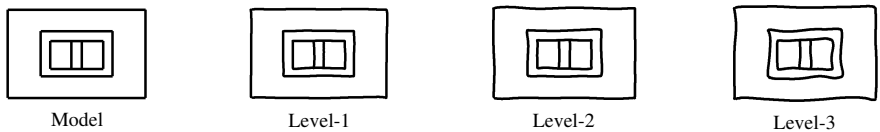

Figure 6. A model symbol with different levels of deformation; used for simulating hand-drawn symbols and applied using an application from project Epéires[3].

---

[2] http://symbcontestgrec05.loria.fr/symboldescription.php (accessed: May 28 2009)

[3] http://www.epeires.loria.fr/

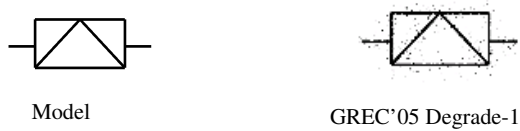

Model                GREC'05 Degrade-1

Figure 7. A model symbol with degraded example; used to simulate photocopying / printing / scanning and applied using ImageMagick[4] and QGar package[5].

Table 1. Results of symbol recognition experiments.

| Number of classes | | 20 | 50 | 75 | 100 |
|---|---|---|---|---|---|
| **Clean symbols** | | 100% | 100% | 100% | 100% |
| Hand-drawn deformation | **Level-1** | 99% | 96% | 93% | 92% |
| | **Level-2** | 98% | 95% | 92% | 90% |
| | **Level-3** | 95% | 77% | 73% | 70% |
| **Binary degrade** | | 98% | 96% | 93% | 92% |

Table 1 summarizes the experimental results. A 100% recognition rate for clean symbols illustrates the invariance of our method to rotation and scaling. The recognition rates decrease with level of deformation and drop drastically for high binary degradations because of irregularities produced in symbol signature; which is a direct outcome of the noise sensitivity of vectorization step. We have not used any sophisticated de-noising or pre-treatment and our method derives its ability to resist against noise, directly from underlying vectorization technique and the fuzzy approach used for computing structural signature. We used only clean symbols for learning and (thus) the recognition rates illustrate the robustness of our system against vectorial and binary noise. The system proposed in [8] presents recognition rates only for 20 models. Figure 8 compares our results with [8].

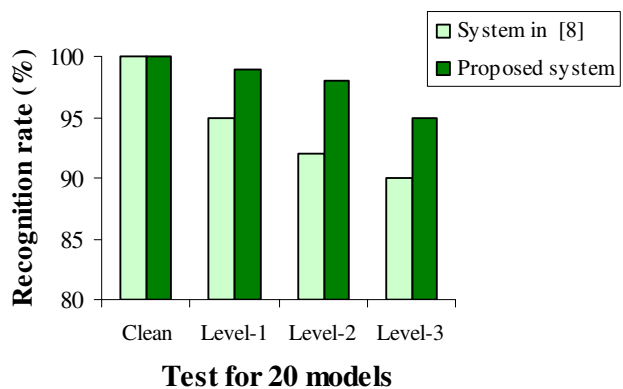

Figure 8. Comparison of recognition rates.

---

## 4.2 Experimentation – symbols with context noise

A second set of experimentation was performed on a synthetically generated corpus of symbols cropped from complete documents [16]. These experiments focused on evaluating the robustness of the proposed system against context noise i.e. the structural noise introduced in symbols when they are cropped from documents. We believe that this type of noise gets very important when we are dealing with complete documents and to the best of our knowledge; we have not found any system that has been evaluated for this type of noise. We have performed these experiments on two subsets of symbols: consisting of 16 models from floor plans and 21 models from electronic diagrams. The models are in Figure 9 and Figure 10. For each class the perfect symbol (model), along with its 36 rotated and 12 scaled examples was used for learning. The examples of models, for learning, were generated using ImageMagick and the test sets were generated synthetically [16] with different levels of context-noise (Figure 11) in order to simulate the cropping of symbols from documents. Test symbols were randomly rotated and scaled and multiple query symbols were included for each class. The test datasets are available at http://mathieu.delalandre.free.fr/projects/sesyd/queries.html (accessed: May 28 2009).

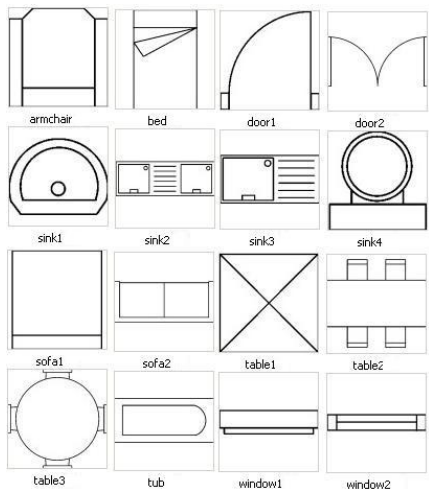
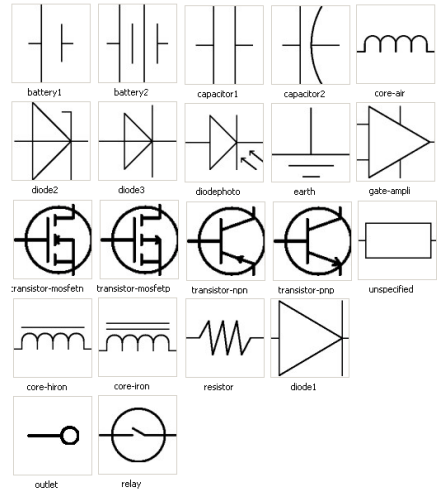

Figure 9. Model symbols from floor plans.     Figure 10. Model symbols from electronic drawings.

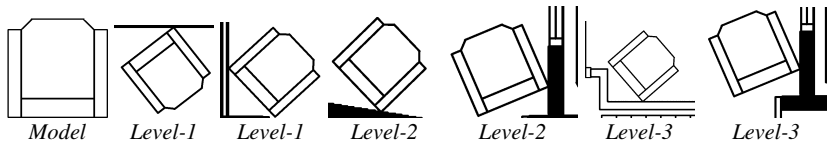

Figure 11. An arm chair with different levels of context noise.

Table 2 summarizes the results of experiments for context noise. We have not used any sophisticated de-noising or pre-treatment and our method derives its ability to resist against context noise, directly from underlying vectorization technique and the fuzzy approach used for computing structural signature. The models for electronic diagrams contain symbols consisting of complex arrangement of lines and arcs, which affects the features in structural signature; as is depicted by the recognition rates for these symbols. But keeping in view the fact that we have used only clean symbols for learning and noisy symbols for testing, we believe that the results show the ability of our signature to exploit the sufficient structural details of symbols and it could be used to discriminate and recognize symbols with context noise.

Table 2. Results of symbol recognition experiments for context noise.

|  | Noise | Model symbol | Query Symbol (each class) | Recog. rate (%) |
|---|---|---|---|---|
| **Floor plans** | Level-1 | 16 | 100 | 84% |
|  | Level-2 | 16 | 100 | 79% |
|  | Level-3 | 16 | 100 | 76% |
| **Average recog. rate** |  |  |  | 80% |
| **Electronic diagrams** | Level-1 | 21 | 100 | 69% |
|  | Level-2 | 21 | 100 | 66% |
|  | Level-3 | 21 | 100 | 61% |
| **Average recog. rate** |  |  |  | 65% |

# 5  Conclusion

Structural methods are the strong methods for graphics representation and statistical classifiers provide efficient recognition techniques. We have presented an original adaptation of Bayesian network learning for the problem of graphic symbol recognition. Our signature exploits the structural details of symbols. We represent symbols by signatures and encode their joint probability distribution by a Bayesian network. We then use Bayesian probabilistic inference on this network to classify query symbols. Experimental results of our method shows an improvement in recognition rates of system in [8] and shows the scalability of the proposed system. Our system does not use any sophisticated de-noising or pre-treatment and it drives its power to resist against deformations and degradations, directly from representation phase. We have addressed the issue of sensitivity of structural representations to noise and deformations; by introducing overlapping intervals for computing structural signature. The features in signature are affected by the small quadrilaterals that are produced during vectorization (in case of noisy symbols), which produce irregularities in signature. The use of fuzzy approach for computing structural signature and probabilistic inference of Bayesian networks gives our system a certain level of resistance against these irregularities. Our experiments have produced encouraging results and we have found that the system is scalable to sufficiently large number of models (classes) but with moderate levels of deformation and degradation. We believe that the recognition rates will be improved for real learning sets which include deformed and degraded examples as well. The system is extensible to new models. The signature is invariant to rotation & scaling and robust against deformations & degradations. It is adapted to underlying symbol set and has a resistance against context noise. The proposed system could be used for 2D linear symbols from a wide range of application domains. The use of lightweight signature and statistical classifier makes our method efficient and scalable to a large number of symbol classes. In future we plan to use this method, as quick graphic symbol discrimination technique, for designing a system for symbol spotting in line drawings.